\documentclass{article}
\usepackage{spconf,amsmath,epsfig}
\usepackage{amsthm,amsmath,amssymb}
\usepackage{mathrsfs}
\usepackage{bm}
\usepackage{multirow}

\let\OLDthebibliography\thebibliography
\renewcommand\thebibliography[1]{
  \OLDthebibliography{#1}
  \setlength{\parskip}{0pt}
  \setlength{\itemsep}{0pt plus 0.3ex}
}

\begin{document}\sloppy

\def\x{{\mathbf x}}
\def\L{{\cal L}}

\title{DSIC: Dynamic Sample-Individualized Connector for Multi-scale Object Detection}
%
\name{Zekun Li\textsuperscript{\rm 1,2*},Yufan Liu\textsuperscript{\rm 1,2*},Bing Li\textsuperscript{\rm 1,3$\dagger$},Weiming Hu\textsuperscript{\rm 1,2,4}}
\address{\textsuperscript{\rm 1}NLPR, Institute of Automation, Chinese Academy of Sciences \\\textsuperscript{\rm 2}School of Artificial Intelligence, UCAS,\textsuperscript{\rm 3}PeopleAI Inc\\\textsuperscript{\rm 4}CAS}

\maketitle

\begin{abstract}
Although object detection has reached a milestone recently, the scale variation is still the key challenge. Integrating multi-level features is presented to alleviate the problems, like Feature Pyramid Network (FPN) and its improvements. However, the specifically designed architectures and fixed data flow paths of these methods are not flexible for feature fusion, especially when fed with various samples. To overcome the limitations, we propose a Dynamic Sample-Individualized Connector (DSIC) for multi-scale object detection, which dynamically adjusts network connections to fit different samples. In particular, DSIC consists of two components: Intra-scale Selection Gate (ISG) and Cross-scale Selection Gate (CSG). With the help of the presented gate operator, ISG adaptively extracts proper multi-level features from backbone as the inputs of feature integration. CSG automatically activates informative data flow paths based on the extracted multi-level features. These two components are both plug-and-play and can be embedded in any backbone. Experimental results demonstrate that the proposed method outperforms the state-of-the-arts.

\end{abstract}
\section{Introduction}
Object detection has been explored for many years as a foundation in computer vision. With the great development of deep learning, object detection has achieved remarkable progress. Plenty of excellent detectors \cite{girshick2014rich,girshick2015fast,ren2015faster,liu2016ssd,redmon2016you,cai2018cascade} are proposed to improve the performance and show extraordinary results on public benchmarks such as MS-COCO \cite{lin2014microsoft}.

\begin{figure}[t]
	\centering
	\vspace{1em}
	\includegraphics[scale=0.5]{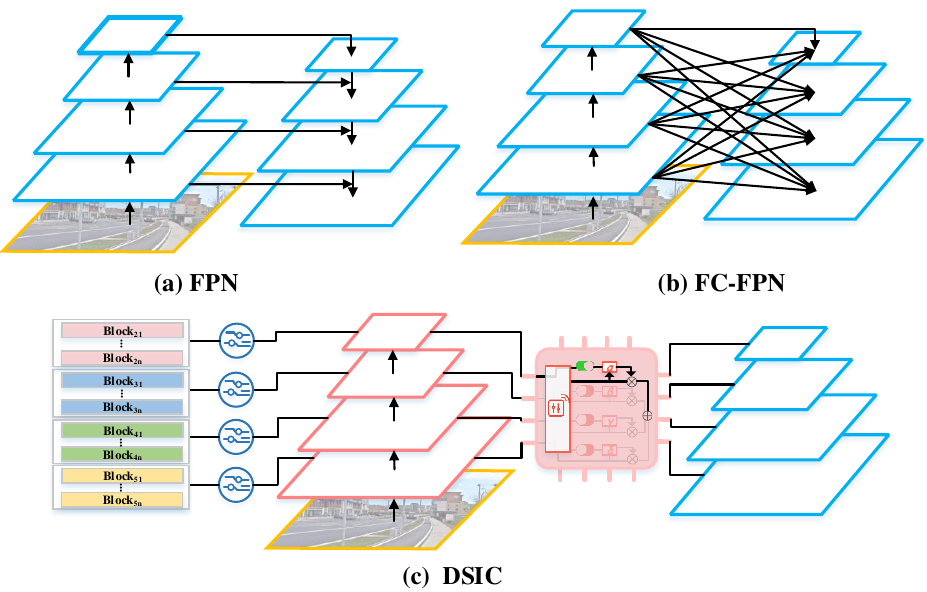}
	\vspace{-0.4em}
	\caption{ Illustrations of (a) the classical FPN, (b) Fully-Connected FPN (FC-FPN) and (c) our method Dynamic Sample-Individualized Connector (DSIC). FPN and FC-FPN represent the different cases of connection which use the same output of backbone as input, constantly. Our DSIC can learn to select input from backbone and activate different data flow paths as connection according to the input sample. }
	\vspace{-1.5em}
	\label{intro}
\end{figure}

However, there are still some problems limiting the performance of detectors. Scale variation is one of the most challenging problems. Feature pyramid Network (FPN) \cite{lin2017feature} is an effective method to alleviate this problem by merging features at adjacent levels to construct a top-down pyramid (seen as Fig \ref{intro} (a)). More recently, some studies \cite{liu2018path,sun2019deep,pang2019libra} have been proposed to improve the connection design in FPN. Nevertheless, the manual designed architectures have fixed connections, ignoring the diversity brought by different samples. And the interaction across multi-level features is not adequate, because the specific connections cannot be the optimal case in various situations. Fully-Connected Feature Pyramid Network (FC-FPN), depicted in Fig \ref{intro} (b), can be seen as the universal set that includes all connection methods between the bottom-up and top-down pyramid. It uses full connection to enhance the feature representation without manual design. But meanwhile, redundancy and noise are brought in, which indicates that it is unnecessary to activate all connections. Furthermore, both series of FPN and FC-FPN are fixed, which are not friendly to different samples. 



In this paper, in order to alleviate aforementioned problems, we propose a novel and effective module called Dynamic Sample-Individualized Connector (DSIC). Different from the methods mentioned above (i.e., Fig 1(a) and (b)), DSIC dynamically selects the proper input and connections, like Fig 1(c) illustrated. It avoids fixed specific design, and automatically adjusts the connection and feature interaction process, to fit different samples. In particular, we firstly present a gate operator as the basic element of DSIC, which controls the state of data flow path. Given control signals, it can connect or disconnect a data flow path, even further enhance or suppress the data flow. Based on the gate operator, DSIC is constructed, consisting of two components: Intra-scale Selection Gate (ISG) and Cross-scale Selection Gate (CSG). ISG aims to explore what is the proper input of feature integration which adaptively extracts multi-level features with sufficient information from backbone. CSG devotes to learning what is the proper connections among multi-level features when fed with different samples. It automatically activates informative data flow paths based on the multi-level features. These two components are both plug-and-play and can be embedded in any backbones separately or jointly. DSIC shows superiority when compared with the state-of-the-arts and other feature integration methods.

In summary, this work makes the following main contributions:
\begin{itemize}
	\item We propose a novel plug-and-play block called DSIC to alleviate the scale-variant problem. To control the connection of data flow path dynamically, we firstly present the gate operator as the basic element of DSIC, which is the core operation of the proposed method.
	\item We propose Intra-scale Selection Gate (ISG) and Cross-scale Selection Gate (CSG) to constitute DSIC, taking advantage of the presented gate operator. ISG aims to adaptively extracts multi-level features from backbone as input, while CSG devotes to automatically activating data flow paths. Each of these components can be separately or jointly embedded into any backbones.
	\item We evaluate the proposed framework on MS-COCO 2017 and it shows the superiority when compared with state-of-the-arts. The ablation experiments validate the effectiveness of each module of DSIC. 
\end{itemize}

\section{Related Works}

\subsection{Multi-level feature extraction}
Scale variation of object instances is a gargantuan obstacle in object detection. The integration of multi-level features is beneficial to mitigate such problem. FPN \cite{lin2017feature} was designed to fuse features through a top-down pyramid. After that, a series of works were presented with improved structures based on FPN, to further enhance the performance. For example, PANet \cite{liu2018path} improved FPN by adding a new bottom-up structure after the feature pyramid to shorten information path. FPG \cite{chen2020feature} was proposed utilizing a deep multi-pathway feature pyramid that repeated the fusion process in different directions. Although the methods above have obtain some progress, there are still some problems. The multi-level features extracted from backbone and data flow paths of integration in these networks are fixed, when different inputs and features are being processed. This is not flexible.

\subsection{Dynamic Mechanisms}
More recently, dynamic mechanisms have been explored to improve the performance of models in computer vision tasks, which adaptively adjust some variables or settings of the network. Some methods use dynamic mechanism to adjust the network configurations. DRConv \cite{chen2020dynamic} assigned filters to learning appointed spatial areas that achieved better performance through obtaining rich and diverse spatial information. Other methods dynamically learn to set the hyper-parameters. Dynamic R-CNN \cite{zhang2020dynamic} was proposed to alleviate the inconsistency between the hyper-parameters and training procedure, by automatically adjusting the IoU threshold and the parameters of loss function. Nevertheless, the existing dynamic mechanisms only focus on training or network configurations adjustment, ignoring the feature settings and data flow path selection. We propose a new dynamic sample-individualized connector that can select the superior features in multi blocks from backbone and the better multi-level feature integration strategy when processing different samples dynamically.

\begin{figure*}[ht]
	\centering
	\vspace{-3.0em}
	\includegraphics[scale=0.75]{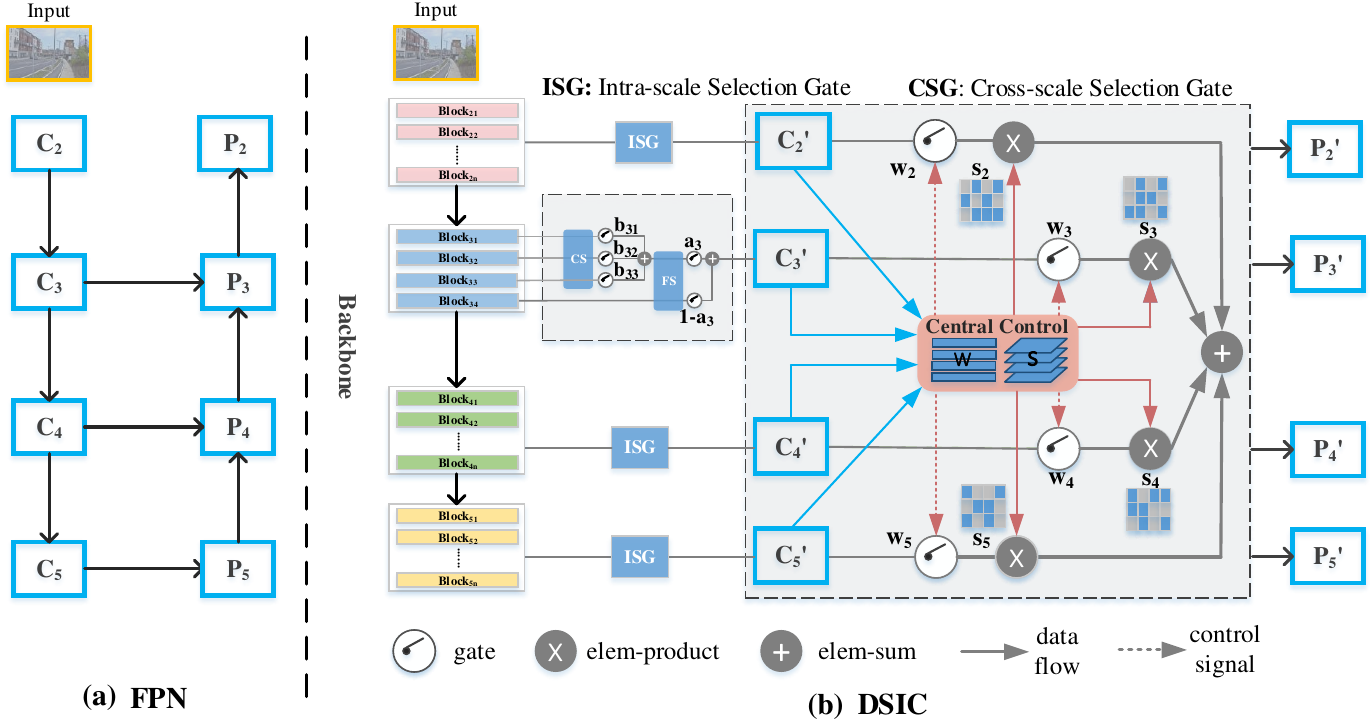}
	\vspace{-1.0em}
	\caption{(a) The conventional feature pyramid network. 
		(b) The framework of the proposed DSIC.}
	\vspace{-2.0em}
	\label{pipe}
\end{figure*}

\section{The Proposed Method}

\subsection{Foundation}
\noindent\textbf{Pipeline}
Existing feature pyramid module is shown in Fig \ref{pipe} (a), which aims to integrating multi-level features to solve multi-scale object detection. In this module,  $\left\lbrace C_2, C_3, C_4, C_5\right\rbrace $ represent the inputs of the feature pyramid. Note that, $C_i$ is extracted from the output of the last block in the $i$-th stage of the backbone. In addition, $ \left\lbrace P_2, P_3, P_4, P_5\right\rbrace $ represent the new feature maps after feature integration. $P_k$ denotes the new feature map at the $k$-th level. Due to the success of alleviating the scale variation problem, feature pyramid module is widely used in the recent years. However, this module has fixed the input and connection data flow path. 

We utilize DSIC to dynamically construct the feature integration module, as illustrated in Fig \ref{pipe}(b). In particular, the proposed DSIC comprises two components: ISG and CSG. ISG is put in the left of the feature integration module, utilized to select the proper features in each stage of the backbone. Note that $Block_{ij}$ denotes the $j$-th block in the $i$-th stage. Besides, CSG is leverage to replace the original connection, to obtain an proper connection cross variant-scale features.

\noindent\textbf{Gate Operator}
The gate operator is the basic element of DSIC, which controls the data flow path. In detail, given data flow $ \textbf{X}  \in \mathbb{R}^{c\times h\times w}$ as input, the gate operator $\mathcal {G}(\cdot)$ can be formulated as:

\begin{equation}
\mathcal{G}(\bm {\varepsilon },\textbf{X})=\sigma[\Psi(\bm \varepsilon, \sum\limits_{\Phi \in \mathcal{F}}\Phi (\textbf{X}))],
\end{equation}
in which $\bm\varepsilon \in \mathbb{R}^{m\times 1\times 1}$ denotes the gate control signal. It assists the gate to determine whether to open or close, and further enhance or suppress the data flow when the gate is open. $\mathcal{F}$  represents the set of functions, including a series of convolutions, making the input compatible with the gate. $\Psi(\cdot)$ denotes the Hadamard product and $\sigma(\cdot)$ represents the mode selection of gate operator by activation function. In this manner, given different data and control signals, the gate shows different states so that the inputs and connection data flow paths are dynamically changed. 

\subsection{Intra-scale Selection Gate (ISG)}
The input with sufficient information is crucial to feature integration. The conventional FPN only extracts the feature maps from each stage's last block as input. This design is rigid, in which only single block's output is used while the information of the rest is lost. Contrastively, we propose ISG to dynamically select the intra-scale information in backbone from coarse to fine and obtain an input with adequate information.


As shown in Fig \ref{pipe} (b), ISG is comprised by Coarse Selection (CS) and Fine Selection (FS). Because of the stronger representation of last block, CS firstly selects the useful information from $(n-1)$ former blocks, to compensate the information that the last block lacks. FS further selects a proper fusion of the last block's output and the complementary data. 

In particular, the $(n-1)$ former blocks $\left\lbrace{B_{ij}}\right\rbrace_{j=1}^{j=n-1}$ in the $i$-th stage are fed into CS, and then the corresponding control signals $b_{ij}$ are obtained to adjust the states of the gates:



\begin{equation} 
\left\lbrace b_{i1},...b_{ij}\right\rbrace=\sum\limits_{\Phi\in \mathcal{I}}\Phi\left\lbrace B_{i1},...B_{ij}\right\rbrace,\quad j=(1,...n-1),
\end{equation}
where $\mathcal{I}$ denotes set of integration operators, including channel-wise concatenation, a series of convolutions and poolings. Note that the parameters of CS in each stage are non-shared. After that, given outputs of blocks and control signals, a series of gates control the data flow paths and obtain:

\vspace{-1.0em}
\begin{equation}
B_{i}^{CS}=\sum_{j=1}^{n-1}\mathcal{G}(b_{ij},B_{ij}) .
\end{equation}
The gate operators $\mathcal{G}(\cdot)$ control whether the data flows from some blocks are needed currently. $B_{i}^{CS}$ denotes the result of selection from previous output. After the process of CS, the results are fed into FS to select an integration of $B_{i,n}$ and $B_i^{CS}$. Similarly, considering $B_{i,n}$ and $B_i^{CS}$, FS computes the gate control signals $a_i$:

\vspace{-1.0em}
\begin{equation}
a_{i}=Max[tanh(F_{gap}(B_{i}^{CS})+F_{gmp}(B_{i}^{CS})),0],   
\end{equation}
where $tanh$ denotes the Tanh activation operator and $F_{gap}$ and $F_{gmp}$ means global average-pooling and global max-pooling, respectively. Here, $a_{i} \in \mathbb{R}^{c\times1\times1}$ is the control signals of $B_{i}^{CS}$ that determines which channels should pass in the gate operator. After the computation of FS, the dynamically selected input of feature integration $C'_i$ can be obtained:

\begin{equation}
C_{i}'=\mathcal{G}(a_{i},B_{i}^{CS})+\mathcal{G}((1-a_{i}),B_{i,n}).  
\end{equation}

Instead of the fixed input of FPN, we rethink the effect of all blocks in backbone and rearrange the retention and abandon of information. The proposed ISG achieves coarse-to-fine selection, which can dynamically extract the useful information as input according to various samples in backbone.

\subsection{Cross-scale Selection Gate (CSG)}
In multi-scale object detectors, feature integration module merges multi-level feature maps $\{C'_2, C'_3, C'_4, C'_5\}$ via lateral connections, in order to obtain a feature pyramid $\{P'_2, P'_3, P'_4, P'_5\}$ with rich semantics at all levels. As shown in Fig 1(b), the full connection in FC-FPN can be seen as the universal set of the lateral connections. This case seems to fully integrate features, but there exists large redundancy. On the other hand, the conventional FPN (seen in Fig 2(a)) is the subset of FC-FPN. It only connects the features at the same level from bottom-up pyramid to top-down pyramid in lateral direction. Both of them are fixed, when fed with different samples.

Different from the cases above, we propose CSG to dynamically activate useful data flow paths, and obtain an flexible connection case when meeting various samples. In detail, as illustrated as Fig 2(b), taking $\{C'_2, C'_3, C'_4, C'_5\}$ as input, the central control unit $\mathcal {CCU(\cdot)}$ firstly generates gate control signals $\textbf{w}_\textbf{k}={\left\lbrace w_{ik}\right\rbrace}_{i=2}^{5}$ and pixel-wise selection maps $\textbf{s}_\textbf{k}={\left\lbrace s_{ik}\right\rbrace}_{i=2}^{5}$ :
\begin{equation} 
\left\lbrace {\textbf{w}}_{\textbf{k}},\textbf{s}_{\textbf{k}} \right\rbrace =\mathcal{CCU}( \left\lbrace\textbf M_\textbf{k} \right\rbrace) \qquad   k=(2,3,4,5),
\end{equation}
%
\begin{equation}
M_{ik}=\left\{
\begin{array}{lr}
F_{down}^{(k-i)}{(C_i')}      & i<k\\
C_i'     &i=k\\  
F_{up}^{(i-k)}{(C_i')}      & i>k\\ 
\end{array}.
\right.
\label{m}
\end{equation}%
Note that $\textbf{M}_\textbf{k}={\left\lbrace M_{ik}\right\rbrace}_{i=2}^5$ are the input feature maps with unified resolution. Specifically, to compute $M_{ik}$, higher-resolution features are down-sampled by the 3$\times$3 convolution and lower-resolution features are
up-sampled by bilinear interpolation with appropriate scale factors. In Equation \ref{m}, $F_{down}^{k-i}$ is 3$\times$3 convolution operators with$ (k-i) $times and $F_{up}^{i-k}$ is bilinear interpolation with scale factor of $(i-k)$.

More detailedly, inside the central control unit, the gate control signals ${\left\lbrace w_{ik}\right \rbrace}_{i=2}^{5}$ are computed as:

\vspace{-0.5em}
\begin{equation} 
{\left\lbrace w_{ik}\right\rbrace}_{i=2}^{5} =\sum\limits_{\Phi\in \mathscr{O}_1}\Phi({\left\lbrace M_{ik}\right\rbrace}_{i=2}^5).
\end{equation}
$\mathscr{O}_1$ indicates the operation set, including convolutions and activation functions. The gate control signals $w_k$ are utilized to adjust the corresponding gate units, so that the useful data flow path can be connected. After the data flow passes the gate, a pixel-wise selection map generated by the central control unit is leveraged to activate the data flow spatially, and further make features from different levels consistent. When it comes to different data flows, the generated selection maps have different activated pixels and significant regions. In particular, inside the central control unit, the pixel-wise selection maps $\textbf{s}_\textbf{k}={\left\lbrace s_{ik}\right\rbrace}_{i=2}^{5}$ at the $k$-th level:


\vspace{-0.5em}
\begin{equation}
{\left\lbrace s_{ik}\right\rbrace}_{i=2}^{5} =\sum\limits_{\Phi\in \mathscr{O}_2}\Phi({{\left\lbrace M_{ik}\right\rbrace}_{i=2}^5})
\end{equation}
$\mathscr{O}_2$ contains a modulus operator that normalize $M_{ik}$ spatially and a series of operations, including convolutions, activation functions and element-wise sum. Subsequently, the final feature pyramid output can be calculated as:
\vspace{-0.5em}
\begin{equation}
{P}_{k}'=\left\lbrack \sum \limits _{i=2}^5(\mathcal{G}( {w}_{ik},{M}_{ik}) \cdot s_{ik}) \right\rbrack,
\end{equation}%
where $P_k'$ denotes the output at the $k$-th level, containing rich information at all levels.

Thus, CSG takes features at all levels into consideration, and automatically selects different connections when meeting different samples. This connection case can eliminate redundant information and achieve a better performance, compared with the full connection case. Besides, the pixel-wise selection can further refine the passed data flow spatially. 

\section{Experiments}

\subsection{Settings}
Our experiments are implemented on MS-COCO 2017 which is a challenging and credible dataset containing 80 object categories. It consists of 115k images for training ($train2017$), 5k images for validation ($val2017$) and 20k test-dev images($testdev$). The training process is performed on $train2017$, and ablation experiments and final results are evaluated on $val2017$ and $testdev$, respectively. The performance is evaluated by standard COCO-style Average Precision (AP) metrics. In order to ensure the fairness of the experiment comparisons, we implement our method and re-implement baseline methods based on PyTorch\cite{paszke2017automatic} and mmdetection\cite{chen2019mmdetection}.



\subsection{Performance}


\textbf{{Comparison with the baseline.}} As shown in Table \ref{base}, DSIC achieves consistent improvement overall all baseline detectors, which brings the definite improvements on various public backbone and different detectors. The better performance proves the generalization and robustness ability of DSIC.  

\begin{table}
	\centering
	\caption{Comparison with baselines on $val2017$. ``$\surd$" means the baseline models integrated with our connector and others mean the baseline models integrated with FPN by default.}
	\vspace{-0em}
	\resizebox{0.48\textwidth }{!}{
		\begin{tabular}{c|c|c|cccccc}
			\hline
			Method&Backbone&DSIC&  $AP$ & $AP_{50}$ & $AP_{75} $&$ AP_{S}$ & $AP_{M}$ & $AP_{L}$ \\
			\hline
			\multirow{4}{*}{FCOS}&\multirow{2}{*}{ResNet-50} & &36.6&55.7&38.8&20.7&40.1&47.4\\
			& &$\surd$&\textbf{37.7}&\textbf{56.5}&\textbf{40.0}&\textbf{22.1}&\textbf{41.2}&\textbf{48.8}  \\
			\cline{2-9}                      
			&\multirow{2}{*}{ResNet-101}&&39.2&\textbf{58.8}&42.1&22.9&42.8&\textbf{51.6 }   \\
			&&$\surd$&\textbf{40.0}&58.4&\textbf{43.1}&\textbf{23.9}&\textbf{43.9}&51.2 \\
			
			\hline
			\multirow{4}{*}{Faster R-CNN}&\multirow{2}{*}{ResNet-50} & &36.3&58.4&39.1&21.5&40.0&46.6\\
			
			& &$\surd$&\textbf{38.3}&\textbf{59.7}&\textbf{41.7}&\textbf{22.5}&\textbf{41.7}&\textbf{49.5}\\
			\cline{2-9}                      
			&\multirow{2}{*}{ResNet-101}&&38.3&60.0&41.8&22.8&42.6&49.5\\
			
			&&$\surd$&\textbf{39.6} & \textbf{61.4} &\textbf{ 43.2} & \textbf{22.8} & \textbf{43.1} & \textbf{50.0} \\
			
			\cline{2-9}

			\hline
			
			\multirow{4}{*}{Mask R-CNN}&\multirow{2}{*}{ResNet-50} & &37.3&59.2&40.4&22.3&40.6&46.3\\
			& &$\surd$&\textbf{38.8}& \textbf{60.5}& \textbf{42.1} & \textbf{22.5}& \textbf{41.8} & \textbf{48.5}\\
			\cline{2-9}
			&\multirow{2}{*}{ResNet-101}&&39.4&60.9&43.1&22.9&43.9&51.1\\
			&&$\surd$&\textbf{40.6}&\textbf{60.7}&\textbf{44.5}&\textbf{24.6}&\textbf{44.4}&\textbf{51.7} \\
			\hline

	\end{tabular}}
	\vspace{-1.5em}
	\label{base}
\end{table}

\noindent\textbf{{Comparison with other feature integration modules.}} As shown in Table \ref{fusion}, we compare our method with other different feature integration modules with same configs. It is obvious that DSIC provides the better performance improvement when compared to common feature pyramid networks including PANet \cite{liu2018path}, FPT \cite{zhang2020feature} and FPG \cite{chen2020feature}. DSIC can dynamically select different connections according to samples avoiding the contradiction between useful and redundant information.
\begin{table}
	\centering
	\caption{Comparison with other feature integration methods on Faster R-CNN with ResNet-50 on $val2017$.}
	\vspace{-0em}
	\resizebox{0.40 \textwidth}{!}{
		\begin{tabular}{ccccccc}
			\hline
			Method &  $AP$ & $AP_{50}$ & $AP_{75} $&$ AP_{S}$ & $AP_{M}$ & $AP_{L}$ \\
			\hline
			FPN&36.3&58.4&39.1&21.5&40.0&46.6\\
			FC-FPN &37.5&59.1&40.3&21.8&41.4&48.7    \\
			PANet & 37.7&59.5&40.7&21.8&41.5&48.9      \\
			FPG & 38.0&59.4&41.2&22.1&40.7&46.4     \\
			FPT & 38.0&57.1&38.9&20.5&38.1&\textbf{55.7}     \\
			Ours& \textbf{38.3}&\textbf{59.7}&\textbf{41.7}&\textbf{22.5}&\textbf{41.7}&49.5    \\
			\hline
	\end{tabular}}
	\vspace{-1.8em}
	\label{fusion}
\end{table}

\noindent\textbf{{Comparison with State-of-the-art.}}
In this section, we evaluate our detector on COCO $test-dev$ set and compare with other state-of-the-art object detection approaches. For a fair comparison, we re-implement the corresponding baselines equipped with FPN on mmdetection. Besides, we use 2x training scheme without any bells and whistles to train our method on Faster R-CNN and Mask R-CNN. All results are shown in Table \ref{sota}. It is obvious that our DSIC boosts the baselines by a significant improvement when integrated with one-stage and two-stage detectors. Also, our method outperforms the state-of-the-art detectors based on the same backbone without any bells and whistles. These improvements demonstrate the superior performance of our proposed framework. 

\begin{table*}
	\vspace{-1.0em}
	\centering
	\caption{Comparisons with the state-of-the-art methods on COCO $test-dev$. The symbol “*” means our re-implemented results on mmdetection.}
	\resizebox{1 \textwidth}{!}{
		\begin{tabular}{ccccccccccc}
			
			\hline
			
			Method  & Backbone & Schedule & $AP$ & $AP_{50}$ & $AP_{75} $&$ AP_{S}$ & $AP_{M}$ & $AP_{L}$ \\
			\hline
			YoLOv2\cite{redmon2017yolo9000}       & DarkNet-19&-   & 21.6 & 44.0 & 19.2 & 5.0 & 22.4 & 35.5      \\
			SSD512\cite{liu2016ssd}       & ResNet-101&-  & 31.2 & 50.4 & 33.3 & 10.2 & 34.5 & 49.8       \\
			Faster R-CNN\cite{ren2015faster}   & ResNet-101-FPN  &- & 36.2 & 59.1 & 39.0 & 18.2 & 39.0 & 48.2      \\
			Deformable R-FCN\cite{dai2016r}   & Inception-ResNet-v2&-   & 37.5 & 58.0 & 40.8 & 19.4 & 40.1 & 52.5 \\
			Mask R-CNN\cite{he2017mask}   & ResNet-101-FPN &-  & 38.2 & 60.3 & 41.7 & 20.1 & 41.1 & 50.2      \\
			Libra R-CNN\cite{pang2019libra}   & ResNet-101-FPN &1x  & 40.3 & 61.3 & 43.9 & 22.9 & 43.1 & 51.0      \\
			\hline
			RetinaNet*   & ResNet-50-FPN  & 1x & 35.8 & 55.6 & 38.4 & 19.8& 38.8 & 45.0      \\
			FCOS*        & ResNet-101-FPN  & 2x & 40.9 & 60.2 & 44.1 & 24.7 & 45.0 & 52.3      \\
			Faster R-CNN*   & ResNet-101-FPN  & 2x & 39.5 & 61.2 & 43.0 & 22.0 & 43.1 & 50.1      \\
			Mask R-CNN*   & ResNet-101-FPN  & 2x & 40.8 & 62.1& 44.6 & 22.8 & 43.9&52.0 \\
			\hline
			DSIC RetinaNet(ours)   & ResNet-50-FPN  & 1x & 37.2 & 57.3 & 39.8 & 20.3 & 40.0 & 47.1   \\
			DSIC FCOS(ours)      	  & ResNet-101&2x   & 41.6 & 60.8 & 44.8 & 25.4 & 45.5 & 53.5\\
			DSIC Faster R-CNN(ours)  & ResNet-101 &2x  & 41.0 & 62.4 & 44.6 & 25.0 & 44.7 & 52.6\\
			DSIC Mask R-CNN(ours)   & ResNet-101 &2x & \textbf{42.6} & \textbf{63.0} & \textbf{46.6} & \textbf{25.5} & \textbf{45.8} & \textbf{52.8}\\
			
			\hline
	\end{tabular}}
	\vspace{-2em}
	\label{sota}
\end{table*}


\subsection{Ablation Studies}
In this section, we conduct the ablation experiments on $val2017$, using Faster R-CNN with FPN based on ResNet-50.

\noindent\textbf{{Ablation studies on each component.}}
In order to verify the importance of our components in DSIC, we apply the ISG and CSG to the model gradually. We report the overall ablation studies in Table \ref{each}. ``CSG" means our method abandons the FPN when compared to the baseline. Both of them have significant improvements and the performance of combination is much better which can prove the effectiveness of association.

\begin{table}
	\centering
	\caption{Effectiveness of each component. ISG: Intra-scale Selection Gate, CSG: Cross-scale Selection Gate.}
	\vspace{-0.em}
	\resizebox{0.48 \textwidth}{!}{
		\begin{tabular}{ccccccccc}
			\hline
			Method& $AP$ & $AP_{50}$ & $AP_{75} $&$ AP_{S}$ & $AP_{M}$ & $AP_{L}$ \\
			\hline
			baseline  &36.3&58.4&39.1&21.5&40.0&46.6\\
			
			baseline+ISG &37.6&59.0&40.5&21.7&40.9&49.0    \\
			baseline+CSG  & 37.8&59.1&41.0&21.8&40.8&48.9      \\

			baseline+ISG+CSG  &  \textbf{38.3}&\textbf{59.7}&\textbf{41.7}&\textbf{22.5}&\textbf{41.7}&\textbf{49.5 }    \\
			\hline
	\end{tabular}}
	\vspace{-1.8em}
	\label{each}
\end{table}

\noindent\textbf{{Ablation studies on Gate operator.}}
We further consider the mode selection in gate operator of two selection gate modules. Experimental results with different modes in two modules are shown in Table \ref{model selection}. 
In ISG, the three modes achieve the similar result which Softmax and Sigmoid ignore the noise in redundant information, which Tanh has a better performance. We consider that the information in the same stage is more complementary instead of mutual exclusion. However, the performance of three modes has obvious differences in CSG because of spatial contradiction of different levels. The Softmax operator takes account of four levels' data flow together ignoring the independence and inconsistency. Sigmoid operator considers the data flow individually which can't eliminate the redundant information. Tanh operator avoids the above defects and achieves the best result. 

\begin{table}
	\centering
	\caption{Comparison with different mode selection of two gate modules.}
	\vspace{-0em}
	\resizebox{0.48 \textwidth}{!}{
		\begin{tabular}{c|c|cccccc}
			\hline
			Module	&	Mode selection&  $AP$ & $AP_{50}$ & $AP_{75} $&$ AP_{S}$ & $AP_{M}$ & $AP_{L}$ \\
			\hline
			\multirow{3}{*}{ISG}&		Softmax& 37.4&58.9&\textbf{40.7}&21.6&41.2&48.2      \\
			&	Sigmoid& 37.5&58.7&40.4&21.7&\textbf{41.3}&47.9     \\
			&	Tanh& \textbf{37.6}&\textbf{59.0}&40.5&\textbf{21.7}&40.9&\textbf{49.0 }    \\
			\hline
			\multirow{3}{*}{CSG}&Softmax& 37.4&59.0&40.4&21.3&41.0&48.4      \\
			&Sigmoid& 37.6&\textbf{59.4}&40.4&\textbf{21.9}&41.4&48.8     \\
			&	Tanh& \textbf{37.8}&59.1&\textbf{41.0}&21.8&\textbf{41.8}&\textbf{48.9}     \\
			\hline
	\end{tabular}}
	\vspace{-2.0em}
	\label{model selection}
\end{table}

\subsection{Visualization}
\begin{figure}[t]
	\centering
	\vspace{1.1em}
	\includegraphics[scale=0.45]{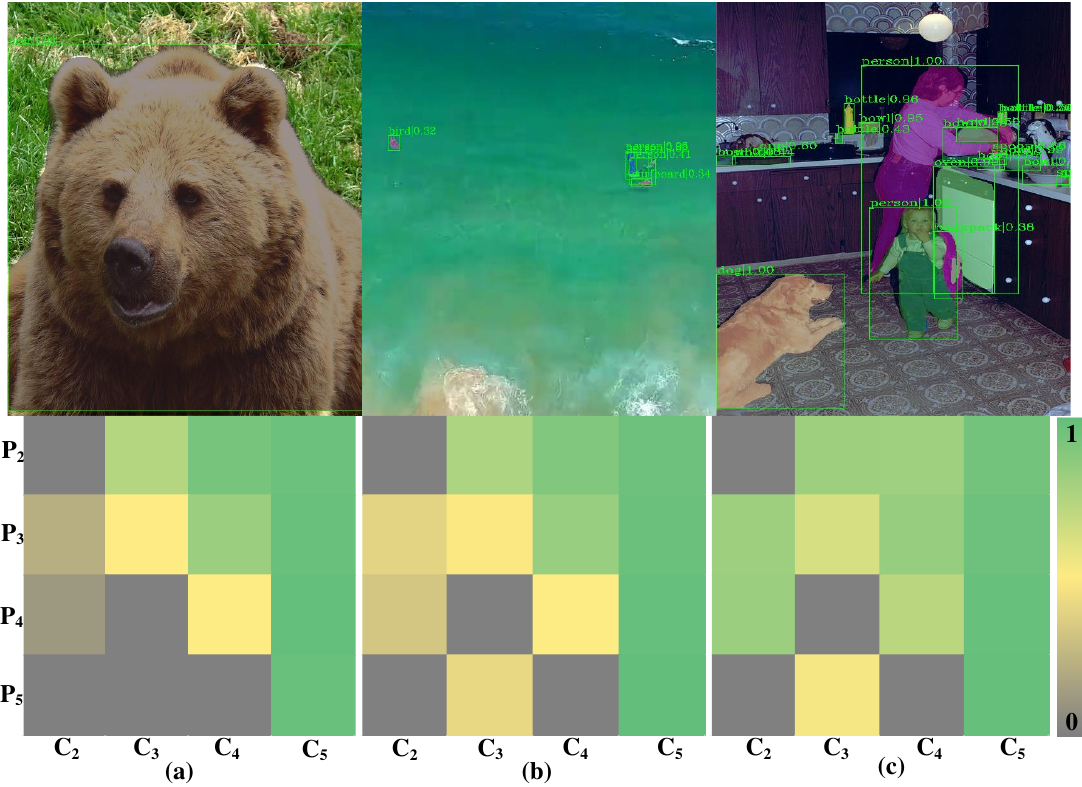}
	\vspace{-1.0em}
	\caption{ The visual results of CSG. The first row is the detection results and the second row is the corresponding state matrix of each data flow path. (a) is the large scale input. (b) is the small scale input. And (c) has various scales of input.}
	\vspace{-1.5em}
	\label{vis}
\end{figure}
We present the visual results of CSG. AS shown in Fig \ref{vis}, three different samples tend to select different connections according to the scale of objects, which validates the sample-individualized data flow path selection of DSIC. In addition, the regression task in detection need more high-resolution information whether large or small scale samples. But we find that the highest-resolution information of output is from the other three levels instead of itself which contains huge noise. By contrast, the classification task need more semantic information in high levels while the highest level output is from itself. Thus, DSIC dynamically selects the inputs and data flow paths of feature integration which reconciles the differences between the classification and regression to some extent.

\section{Conclusion}
In this paper, we have proposed a novel dynamic sample-individualized connector (DSIC) for multi-scale object detection, which dynamically adjusts network connections to fit different samples. With the help of two simple yet effective components, i.e., ISG and CSG, DSIC has shown the generality and effectiveness for both two-stage and single-stage detectors. Compared with previous approaches, DSIC offers several advantages: (1) instead of utilizing the special manual design and insufficient interactions, the whole process is dynamic and sample-individualized; (2) it performs better on multi-level feature integration and can be simply generalized in different computer vision tasks. The experiments on MS-COCO show the superiority of DSIC.

\small
\bibliographystyle{IEEEbib}
\bibliography{myreficme}

\end{document}